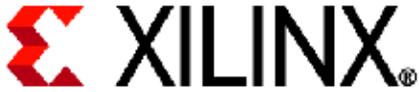

**WP537 (v1.0)**

# Adaptive Computing in Robotics
## Leveraging ROS 2 to Enable Software-Defined Hardware for FPGAs

By: Víctor Mayoral-Vilches and Giulio Corradi

*Xilinx's adaptive SOMs are the perfect compute platform for robotics, allowing the creation of software-defined hardware for robots, delivering solutions with increased performance per watt while also being cost-effective, energy efficient, secure, safe, and adaptable.*

**ABSTRACT**

Traditional software development in robotics is about programming functionality in the CPU of a given robot with a pre-defined architecture and constraints. With adaptive computing, instead, building a robotic behavior is about programming an architecture. By leveraging adaptive computing, roboticists can adapt one or more of the properties of its computing systems (e.g., its determinism, power consumption, security posture, or throughput) at run time.

Roboticists are not, however, hardware engineers, and embedded expertise is scarce among them. This white paper adopts a ROS 2 roboticist-centric view for adaptive computing and proposes an architecture to include FPGAs as a first-class participant of the ROS 2 ecosystem. The architecture proposed is platform- and technology-agnostic, and is easily portable. The core components of the architecture are disclosed under an Apache 2.0 license, paving the way for roboticists to leverage adaptive computing and create software-defined hardware.

# Introduction

A robot is a system of systems—one that comprises sensors to perceive its environment, actuators to act on it, and computation to process it all, while responding coherently to its application. To a large extent, robotics is the art of system integration [Ref 1], both in terms of software and hardware. Past studies show that up to 70% [Ref 1] [Ref 2] of the resources in robotics are dedicated to integration, as opposed to the development of the end application. It is only lately, with the wide availability of lower-end industrial robots, that companies are building on top of hardware, focusing solely on software. Still, robots are highly specialized systems designed to carry out a set of tasks with high reliability and precision. As a result, the relationship between hardware and software capabilities in a robot is critical.

Most robots exchange information across their internal networks while meeting timing deadlines. In a way, a robot is a network of time sensitive networks.

Robotic systems usually have limited onboard resources, including memory, I/O, and disk or compute capabilities, which impede the system integration process and make it difficult to meet real-time requirements [Ref 3] in unstructured, dynamic, or changing environments. This is true with the advent of cybersecurity in robotics [Ref 2] [Ref 4] [Ref 5] [Ref 6] [Ref 7] [Ref 8] [Ref 9], which often imposes new requirements during the lifetime of a robot that demand changes in its logic, affecting real-time loops [Ref 10] [Ref 11] [Ref 12] [Ref 13]. It is, therefore, essential to choose a proper compute platform for the robotic system—one that simplifies system integration, meets power restrictions, and adapts [Ref 14] [Ref 15] to the changing demands of robotic applications.

Adaptive robots are those with the ability to respond successfully to a new situation. For a robot to be considered adaptive at least one of the three cornerstones must be met. Robots meeting all the cornerstones can be considered fully adaptive robots. See Figure 1.

| Cornerstone 2 | **Adaptive Behaviors**<br>(Ref 1, 16, 17, 18, 19, 20, 21, 22) |
|---|---|
| Cornerstone 1 | **Adaptive Mechatronics**<br>(Ref 23, 24, 25, 26, 27, 28, 29, 30, 31) |
| Cornerstone 0 | **Adaptive Computing**<br>(Ref 3, 10, 13, 32, 33, 34, 35, 36, 37, 38, 39) |

WP537_01_080921

*Figure 1:* **Adaptive Robotics Cornerstones**

***Key Message:*** *Robots are highly specialized systems designed to carry a set of tasks with high reliability and precision. There is a critical relationship between the hardware and the software capabilities in a robot. It is thereby essential to choose a proper compute platform for the robotic system. One that simplifies system integration, meets power restrictions and adapts to the changing demands of robotic applications.*

Adaptive mechatronics is a concept that has been around for a few decades. Gosselin [Ref 23] discussed the topic from a mechanical point of view and defined a system to be adaptive if it has a capacity to respond successfully to a new situation. He also defined an adaptive robotic mechanical system as an adaptive system when the ability to adapt to new external situations relies strictly on mechanical properties. Gosselin provided a variety of early examples of adaptive robotic systems that relied purely on mechanical constructs, including adaptive robotic hands [Ref 24] [Ref 25] [Ref 26]. Ivanov [Ref 27] also studied adaptive robotics from a pure mechanical point of view, wherein he proposed how adaptive robots should consider using adaptive electric drives with the properties to change their output speed depending on the payload. He named such adaptive conduct *self-regulation* and argued that adaptive electric drives provided high power efficiency in the robot. Extending this work into mechatronics through control mechanisms using sensory input, there is research on adaptive robot control with sensory feedback. Examples include visual feedback [Ref 28] [Ref 29] or force sensors feedback [Ref 30] [Ref 31], among many others.

Adaptive behaviors in robotics is not a novel concept. It can be traced back to Brooks behavior-based robotics approach and his subsumption architecture in the mid-1980s [Ref 16] [Ref 17]. During the 1990s, various groups researched how to make robots more autonomous by providing them with the capacity for adaptation and self-organization, often with the control mechanism using some form of artificial neural networks connecting the robot's sensors and actuators. This was later summarized by Ziemke [Ref 18] with the concept of adaptive neuro-robotics, which coined one of the first uses of the term *adaptive robotics* and referred to the use of artificial neural systems and adaptive techniques for the control of autonomous agents. More recently, in the robotics compilation in *Robotic Fabrication in Architecture, Art and Design 2018* [Ref 19], the authors refer to adaptive robotics to illustrate solely adaptive behaviors (as opposed to "Adaptive Robotics" [Ref 27], which mostly focuses on mechanical constructs), by means of additional sensing and processing. According to the manuscript, adaptive robots are those with the capacity to adapt to changing environmental conditions and material features by means of additional sensors, while retaining a degree of predictability. Contemporary to this and on the same line of thought, Mayoral-Vilches et al. [Ref 20] [Ref 21] introduced the concept of self-adaptable robots, which leveraged hardware modularity [Ref 1] and Artificial Intelligence [Ref 22] (referring to the adaptive neuro-robotics trend in the 1990s) to reduce the effort and time required to build such robots.

Adaptive computing is the third cornerstone, and in the context of robotics it refers to the capability of a robot to adapt one or more of the properties of its computing systems (e.g., its determinism, power consumption, or throughput) during run time. As described in "What is adaptive computing?" [Ref 32], FPGAs are an ideal technology to realize adaptive computing. FPGAs were first introduced by Freeman [Ref 40] in 1984, laying the foundation of adaptive computing. They are versatile and powerful while also being efficient and cost-effective. FPGAs can be used for virtually any processing task in robotics since it is possible to implement parallel processing on an FPGA in addition to implementing other processing architectures. Another aspect that makes FPGAs ideal for adaptive computing is that datapath widths and register lengths can be tailored specifically to the needs of every robotics application. Examples of how adaptive computing is used in robotics include the design of compute pipelines to accelerate motion planning [Ref 33] [Ref 34] [Ref 35] [Ref 36] [Ref 37] [Ref 41], localization [Ref 38] [Ref 39], distributed synchronization [Ref 13], or time-sensitive resilient communications [Ref 10], among others. More examples of adaptive computing in robotics can be found in "A survey of FPGA-based robotic computing" [Ref 3] and "A survey on FPGA-based sensor systems: towards intelligent and reconfigurable low-power sensors for computer vision, control and signal processing." [Ref 42]

This white paper aims to lay the foundation of adaptive computing in robotics and proposes a software architecture to generate ROS 2 software-defined hardware. Adaptive Computing for Robots introduces adaptive computing for robots providing an overview of the core technologies and capabilities of adaptive computing, explaining why FPGAs are relevant in robotics and connecting to the concept of software-defined hardware. Bringing Adaptive Computing to ROS 2 provides a quick overview of ROS 2 and past related work on adaptive computing. Then, it describes the architecture proposed to bring adaptive computing to ROS 2 as a first class participant. Finally, the Conclusion summarizes the most important points.

# Adaptive Computing for Robots

## Compute Technologies in the Robotics Context

CPUs and general-purpose GPUs (GPGPUs) are the two widely used commercial compute platforms due to their availability and generalized use [Ref 43]. The general-purpose nature of these compute technologies make them especially interesting for roboticists, however, this comes at a cost:

1. Their fixed architectures have difficulty adapting to new robotic scenarios. Additional capabilities often require additional hardware, which usually implies additional system integration.
2. Their general-purpose nature leads them to time inefficiencies, which impact determinism (difficult to meet hard real-time deadlines).
3. Their power consumption is generally one or two orders of magnitude above[1] specialized compute architectures (e.g., through FPGAs or ASICs).
4. Their fixed and non-adaptive architectures make them more vulnerable to cybersecurity threats and malicious actors. Example cyberattacks, such as Meltdown [Ref 44] or Spectre [Ref 45], showed how the lack of capabilities to reconfigure dataflow pipelines leads to eventually insecure compute platforms.

***Key Message:*** *The fixed architectures of CPUs, GPUs, and ASICs come at a cost. Their difficult adaptability leads to time inefficiencies, consumes more power, and makes them more vulnerable to cyber-threats since they cannot reconfigure themselves to apply hardware mitigations.*

---

1. According to [Ref 3], a typical CPU can achieve 10–100 GFLOPS with below 1GOP/J power efficiency. In contrast, GPU is designed with thousands of processor cores running simultaneously, which enables massive parallelism. A typical GPU can perform up to 10 TOPS performance and becomes a good candidate for high-performance scenarios. Recently, benefiting in part from the better accessibility provided by CUDA/OpenCL, GPUs have been predominantly used in many robotic applications. However, conventional CPUs and GPUs usually consume 10W to 100W of power.

### *Industrial Analogy for CPUs*

The industrial analogy of Figure 2 depicts how a CPU can be understood as a group of workshops, with each one employing a very skilled worker.[1]

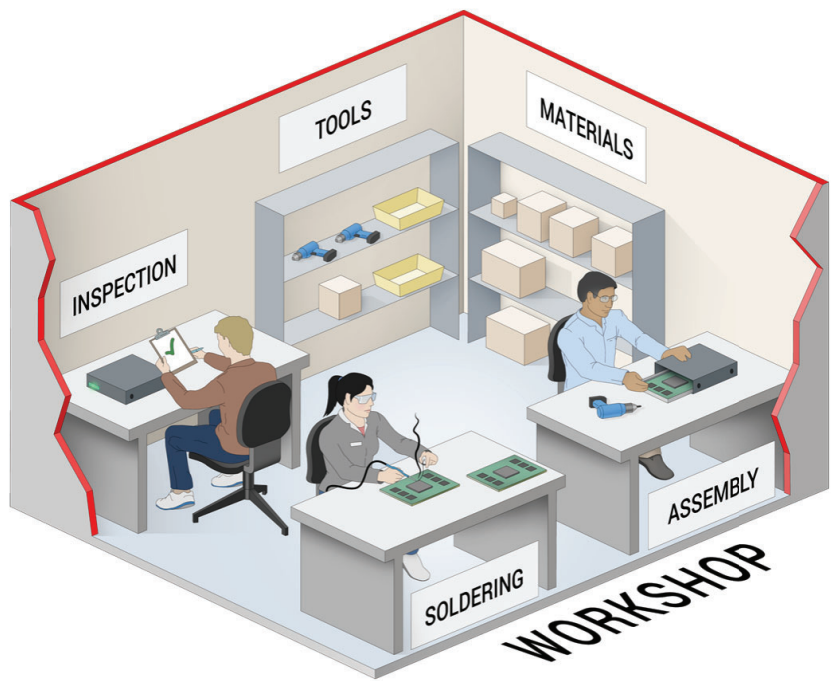

*Figure 2:* **Industrial Analogy for CPUs**

These workers each have access to general-purpose tools that let them build almost anything. Each worker crafts one item at a time, successively using different tools to turn raw material into finished goods. This sequential transformation process can require many steps, depending on the nature of the task. The workshops are mostly (disregarding the cache) independent, and the workers can all be doing different tasks without distractions or coordination problems. Although CPUs are highly flexible, their underlying hardware is fixed. Most CPUs still operate on the basic Von-Neumann architecture [Ref 47] (or more accurately, stored-program computer), where data is brought to the processor from memory, operated on, and then written back out to memory. Fundamentally, each CPU operates in a sequential fashion, one instruction at a time, and the architecture is centered around the arithmetic logic unit (ALU), which requires data to be moved in and out of it for every operation [Ref 48].

---

1. For additional context on the industrial analogy of Figure 2, the reader is referred to [Ref 46].

*Industrial Analogy for GPUs*

A GPU also has workshops and workers, but it has considerably more of them, and the workers are much more specialized as depicted in Figure 3.

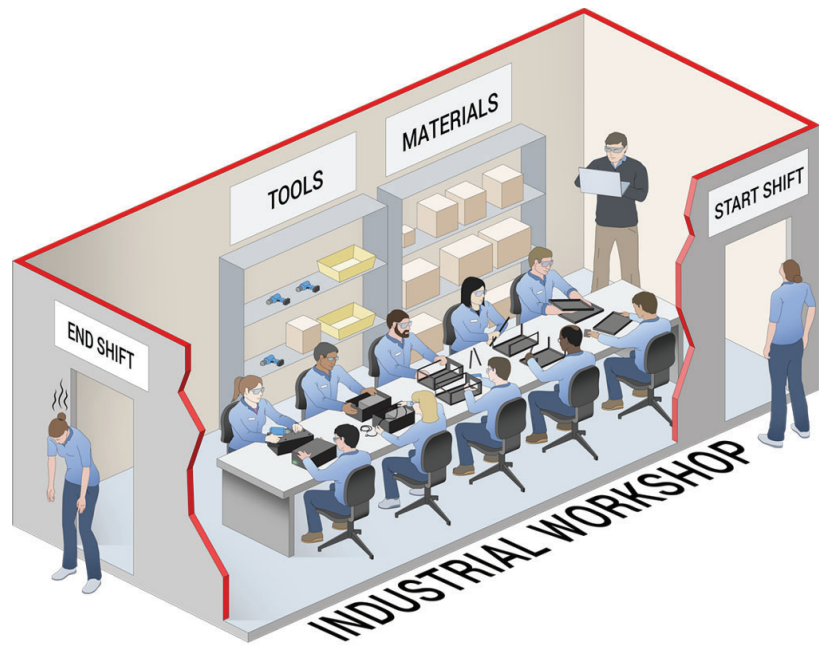

*Figure 3:* **Industrial Analogy for GPUs**

These workers have access to only specific tools and can do fewer things, but they do them very efficiently. GPU workers function best when they do the same few tasks repeatedly, and when all of them are doing the same thing at the same time. After all, with so many different workers, it is more efficient to give them all the same orders. GPUs address one of the major drawbacks of CPUs—the ability to process large amounts of data in parallel.

Though with a significantly higher number of cores when compared to a CPU, GPUs also have a fixed hardware architecture. The core of a GPU still contains a type of a Von-Neumann processor. A single instruction can process a thousand pieces of data or more, though, typically, the same operation must be performed on every data point being simultaneously processed. The atomic processing elements operate on a data vector (as opposed to a data point in the case of CPUs), but still perform one fixed instruction per ALU. Data, thereby, must also still be brought to these processing units from memory via a fixed datapath [Ref 48]. Like CPUs, GPUs are built with fixed hardware—the basic architecture and dataflow is fixed for all robotic applications.

### Industrial Analogy for FPGAs

If CPUs and GPUs are workshops with workers taking sequential steps to transform inputs into outputs, FPGAs are flexible and adaptable factories with assembly lines and conveyor belts custom created for the particular task at hand (see Figure 4).

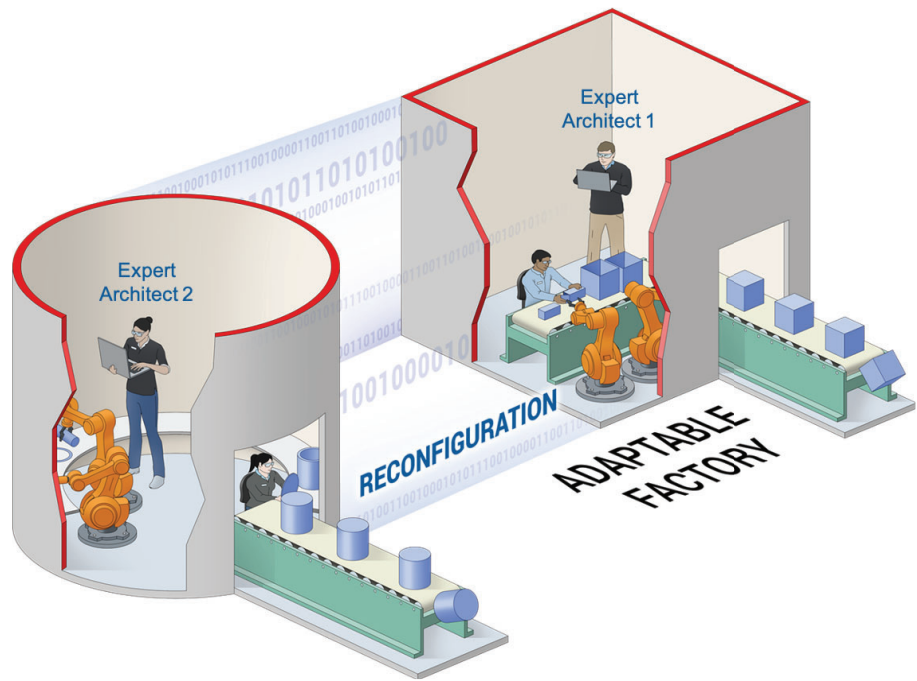

Figure 4: **Industrial Analogy for FPGAs**

This adaptability means that the FPGA architects get to build factories, assembly lines, and workstations, and then customize them for the required task instead of using general-purpose tools. Raw materials within these factories are progressively transformed into finished goods by groups of workers dispatched along assembly lines. Each worker performs the same task repeatedly, and the partially finished product is transferred from worker to worker on the conveyor belt. This results in a much higher production throughput and an optimal use of resources and power. In this analogy, factories are OpenCL acceleration kernels, assembly lines are dataflow pipelines, and workstations are OpenCL compute functions.

*Industrial Analogy for ASICs*

Like FPGAs, ASICs build factories, however, the ones in an ASIC are final and cannot be modified (see Figure 5). In other words, these ASICs contain only robots, and no human cognition is present in the factory. The assembly lines and conveyor belts are fixed, allowing no changes in the automation processes. The ad hoc fixed architectures of ASICs allow them to offer unmatched power efficiency as well as the best prices for high-volume mass production. Unfortunately, ASICs take many years to develop, and do not allow for any changes, which rapidly leads to assets that cannot keep up with future productivity enhancements.

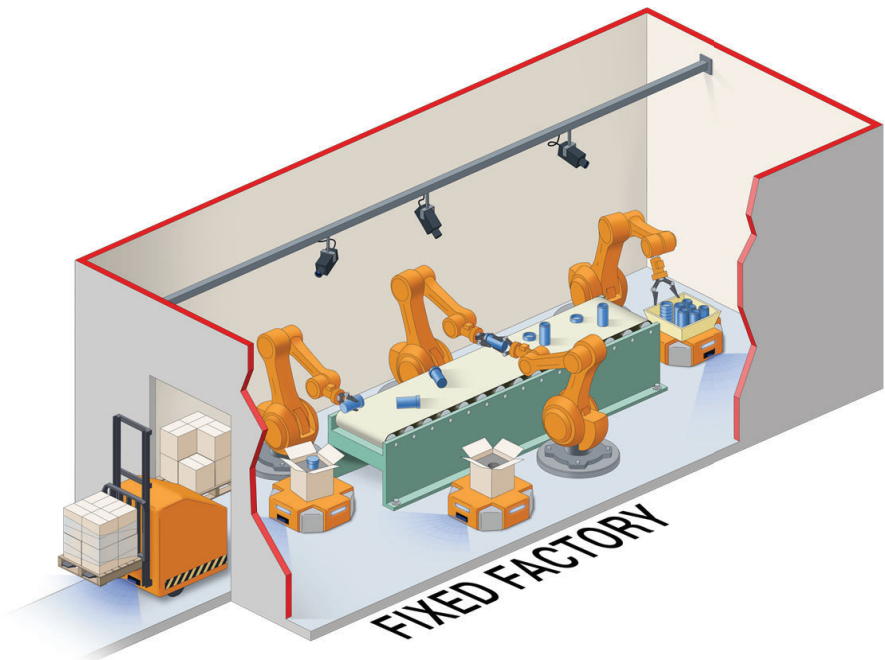

*Figure 5:* **Industrial Analogy for ASICs**

## Why FPGAs Matter in Robotics

CPUs and GPUs excel in control flow computations. Their control-driven machine model is based on a token of control, which indicates when a statement should be executed. This gives CPUs and GPUs full control to easily implement complex data and control structures, however, this also comes at the cost of being less efficient and difficult to program accurately (error-free). FPGAs instead excel in dataflow computations. They follow a data-driven machine model, and statements are executed as soon as all operands are available. This leads to a high potential for parallelism and throughput, while remaining free from errors or side effects.

Overall, presented as an alternative to CPU and GPU general-purpose platforms, FPGAs can adaptively generate custom computing architectures to meet the robotic demands. They are heavily used by popular robot industrial manufacturers as well as in healthcare robotic applications because of their unprecedented flexibility and reduced design cycle and development cost [Ref 43] [Ref 45] [Ref 49] [Ref 50] [Ref 51] [Ref 52]. In [Ref 3], the reader can find a survey of FPGA-based robotic computing, illustrating how FPGAs fits well in robotics applications. Following are the features in detail:

- **Adaptive:** FPGAs can generate unmatched custom computing architectures when there are hard real-time and mixed-criticality requirements with a combination of control flow and dataflow that is prohibitive for CPUs and GPUs because of latency and response time issues. This is due to the CPUs and GPUs having fixed computer architectures, which limit their overall capabilities, including response time and latency.

- **High Performance:** FPGAs increase the computational performance by creating deeply pipelined datapaths (streaming computing), rather than multiplying the number of compute units like CPUs and GPUs. Streaming computing works with the principle that data produced by a computation unit is immediately processed by the next computational unit in the pipeline. This removes fetch-compute-store approaches from the dataflow pipelines in favor of compute producer and consumer operations, increasing performance. Instead, CPUs and GPUs address computations in a suboptimal manner performance-wise, presenting limitations due to their fixed architectures, with a fix number of cores, a fixed instruction set, and a rigid memory architecture.

- **Power Efficient:** CPUs and GPUs address computations in a suboptimal manner (performance-wise), presenting performance limitations due to their fixed architectures, with a fix number of cores, a fixed instruction set, and a rigid memory architecture.

- **No Wasted Computation:** FPGAs trade flexibility against silicon real estate to gain performance. Dynamic Function eXchange (DFX—previously called partial reconfiguration) allows a threaded application running on a CPU to time-share the FPGA such that while a given thread is consuming the result produced by the FPGA, another thread can use the FPGA for a different computation.

- **Predictable:** FPGAs help CPUs and GPUs in offloading strict real-time computations, allowing nanosecond predictability in the execution time and independence from software changes, or jitter, associated with GPU and CPU computation.

- **Reconfigurable:** Robotic algorithms are still evolving rapidly, and FPGAs can be dynamically reconfigured and updated as needed. Moreover, FPGAs can easily be re-programed to meet heterogeneous needs, obtaining the general-purpose characteristics that CPUs and GPUs often strive for.

- **Secure:** FPGAs provide flexibility to build security circuitry on-demand to secure robotic flows. In addition, FPGAs can leverage reconfiguration to patch flaws on its hardware architectures (hardware mitigations). This enables designers to quickly address security hazards that might be difficult or impossible to address on fixed compute architectures (mitigate future vulnerabilities like Meltdown [Ref 44] and Spectre [Ref 45]).

Arguably [Ref 3], FPGAs are great compute substrates for roboticists, however, the flexibility they offer can come at the cost of additional complexity and required design skills. "A survey of FPGA-based robotic computing," [Ref 3] lists some of the additional skills required. The best results in robotics are obtained when one combines all the technologies together holistically: CPUs with many cores and multi-cores, GPUs, and FPGAs. Xilinx provides such integrated System-on-Chip (SoC) solutions that combine the general-purpose software programmability of a CPU with the adaptive hardware capabilities of an FPGA, all in the same device. By doing so, these adaptive SoCs provide a dual software and hardware flexible compute infrastructure for robotic applications, delivering high performance, low power, deterministic, hardware reconfigurable, secure, and adaptive characteristics.

***Key Message:*** *CPUs and GPUs excel in control flow computations, and FPGAs excel in dataflow computation. Adaptive SoC solutions provide a dual software and hardware flexible compute substrate for robotic applications, delivering low power, high performance, deterministic, hardware reconfigurable, secure, and adaptive characteristics.*

## Software-Defined Hardware for Robots

The term *software-defined hardware* often refers to the mapping of applications to an FPGA, which enables the creation of runtime-reconfigurable hardware through software. Software-defined hardware aims to maximize the run-time efficiency of specific algorithms or computations and is an alternative to the fixed Von-Neumann based compute architectures offered by CPUs and GPUs, or the expensive and immutable ASICs. Software-defined hardware for robots should, therefore, be understood as runtime-reconfigurable robot hardware that can be re-programmed and adapted via software.

Traditional software development in robotics is about programming functionality in the CPU of a given robot with a pre-defined architecture and constraints. As described earlier, this leads to complex system integration efforts whenever the robot encounters the need for adaptation. With FPGAs, instead, building a robotic behavior is about programming an architecture that solves the task. Robotic architects have then the possibility to create their own hardware designs purely from software and deliver it through various platforms as illustrated in Figure 6.

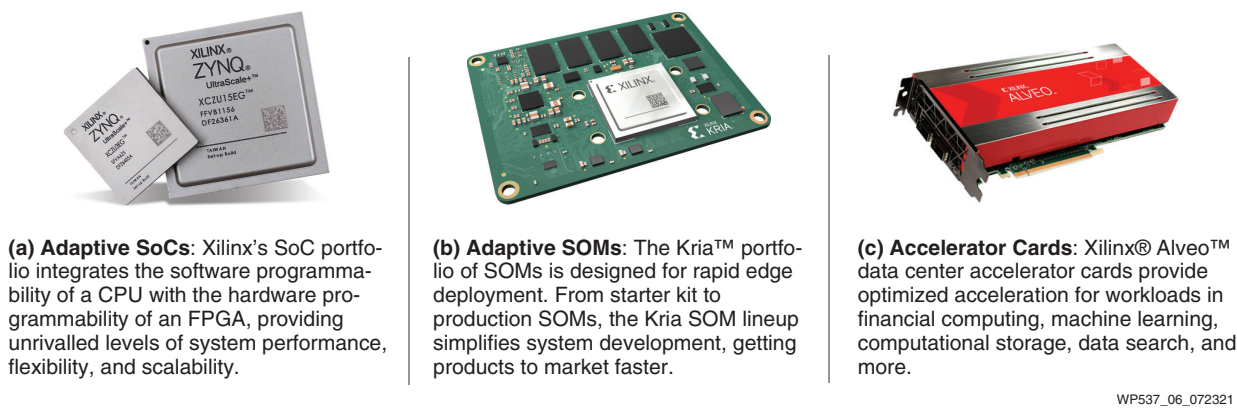

**(a) Adaptive SoCs**: Xilinx's SoC portfolio integrates the software programmability of a CPU with the hardware programmability of an FPGA, providing unrivalled levels of system performance, flexibility, and scalability.

**(b) Adaptive SOMs**: The Kria™ portfolio of SOMs is designed for rapid edge deployment. From starter kit to production SOMs, the Kria SOM lineup simplifies system development, getting products to market faster.

**(c) Accelerator Cards**: Xilinx® Alveo™ data center accelerator cards provide optimized acceleration for workloads in financial computing, machine learning, computational storage, data search, and more.

WP537_06_072321

*Figure 6:* **Xilinx Adaptive Computing Solutions**

For roboticists, there are three ways to interact with FPGA technology. First, a chip-down approach (Figure 6a), where the System-on-Chip (SoC) is integrated into a custom-designed PCB to meet the application needs. This approach is best for robot manufacturers and ideal for large and cost-optimized batches. The second approach is with System-on-Modules (SOMs) (Figure 6b), pre-assembled boards pluggable into a custom carrier board. SOMs help hardware engineers build products faster, abstracting themselves from the compute platform and focusing on the added value. The third approach is with a fully assembled board (Figure 6c), in which many of the peripherals are included. For computationally heavy operations, a full board pluggable in a workstation represents the best trade-off.

*Key Message:* *Traditional software development in robotics is about programming functionality in the CPU of a given robot with a pre-defined architecture and constraints. With adaptive computing, building a robotic behavior is about programming an architecture.*

# Bringing Adaptive Computing to ROS 2

## Why ROS 2?

The Robot Operating System (ROS) [Ref 53] is the de facto framework for robot application development. Maintained and steered by Open Robotics, ROS is not an operating system, but a framework. It consists of different tools to build and manage robots including debugging and visualization utilities, orchestration tools, robotics libraries (e.g., motion planning, navigation, localization, etc.), and communication piping meant to facilitate the development of robotic systems.

Currently, the original ROS article has been cited about 8,500 times, which shows its wide acceptance for research and academic purposes. ROS was born in this environment; its primary goal was to provide the software tools that users need to undertake novel research and development. ROS's popularity has continued to grow in industry, supported by projects like ROS-Industrial (ROS-I for short), an open-source initiative that extends the advanced capabilities of ROS software to industrial applications. Spearheaded by the ROS-Industrial consortium, ROS deployment in industry is now a reality. To date, the consortium has more than 80 members, and its gatherings in Europe, USA, and Asia bring together hundreds of robotics experts every year with the presence of popular manufacturers introducing their own ROS drivers in recent demonstrations.

As ROS outgrew academy and started being used in industry and other domains, the limitations of ROS became increasingly obvious: lack of embedded and deep embedded native support, single-robot software architecture, no real-time capabilities, and lack of security, among others. To address these issues, Open Robotics started redesigning ROS in 2014, which led to the creation of ROS 2. ROS 2 addresses most of the identified limitations by partitioning the communication middleware from the robotics logic. Particularly, Open Robotics selected Data Distribution Service (DDS) as the initial communication middleware and built adapters for various DDS solutions while exposing DDS features to upper layers. Still, ROS core layers remain communication middleware agnostic. The ROS 2 software architecture is depicted in Figure 7:

```
+-----------------------------------------------+
|                   user land                   |
+-----------------------------------------------+
|                     tools                     |
+-----------------------------------------------+
|           ROS client library (rcl)            |
+-----------------------------------------------+
|         ROS middleware interface (rmw)        |
+---------------+---------------+---------------+
| DDS adapter 1 | DDS adapter 2 | Other middle. |
+---------------+---------------+---------------+
|   DDS impl 1  |   DDS impl 2  |  Other impl 3 |
+---------------+---------------+---------------+
                                      WP537_07_082521
```

*Figure 7:* **ROS 2 Software Architecture**

The upper layers build on top of a middleware abstraction layer (rmw), which translates ROS abstractions into middleware specific ones. The ROS client library (rcl) does not expose any middleware-specific implementation details (e.g., DDS). This way, rcl remains middleware agnostic and can be easily extended to other applications requiring a different transport.

With thousands of active users, ROS is by far the biggest community of roboticists. Born out of research, it has evolved for more than a decade, used across areas of application, and is able to serve industrial needs.

## Understanding ROS 2 Concepts

ROS provides tools, libraries, and conventions along with a growing community of roboticists. Conceptually, most aspects of ROS revolve around an abstraction called the ROS computational graph. Each node within the computational graph performs a robotic computation and exchanges information with other nodes through a common peer-to-peer databus, implemented by the underlying communication middleware. Channels of communication within the databus are organized by topic. The overall robotic behavior is, therefore, determined by the computational graph, which can be implemented in one or multiple computers (in a distributed manner). This leads to a second abstraction that maps the computational graph to the compute substrates available in the robot, the ROS data layer graph. The data layer graph represents the physical groupings and connections that implement the behavior modeled in the computational graph. Simply put, it captures the physical reality of the robot including the communication buses, the robot components (including sensors and/or actuators), and the mapping of the computational graph to the compute substrates available in the existing robot components.

ROS computational graphs can involve one or multiple robots and are by nature modular, with the possibility of being distributed or centralized. Figure 8 depicts both the ROS computational graph (the top portion of Figure 8) and the data layer graph (the bottom portion of Figure 8). Table 1 summarizes some of the most important ROS concepts.

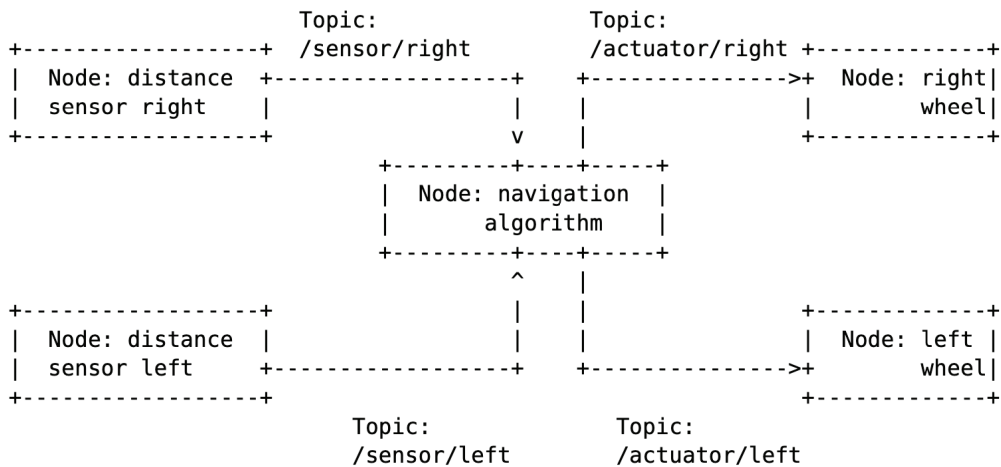

(a) **ROS 2 computational graph**: A data structure that models the overall robotic behavior through each individual computation represented as a *Node*.

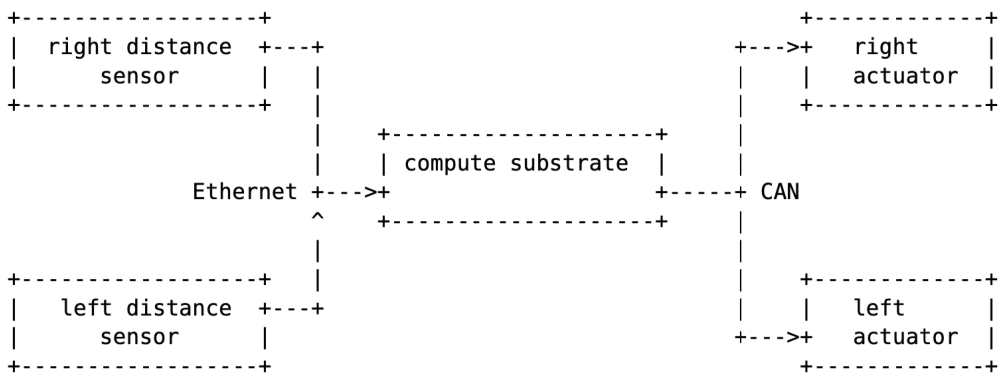

(b) **ROS 2 data layer graph**: Represents the physical groupings and connections, which implement the behavior modeled in the computational graph.

WP537_08_072521

*Figure 8:* **ROS Abstractions for a 2-Wheeled Robot with Navigation Capabilities**

*Table 1:* **Summary of the Most Relevant ROS Concepts**

| ROS concept | Interpretation |
|---|---|
| Computational Graph | A data structure that models the overall robotic behavior through each individual computation represented as a node, communicating with other computation nodes through topics and other abstractions. The computational graph not only helps visualize the robotic behavior but also drives the design process by partitioning each robotic computation into nodes. From an electrical engineer's perspective, the computational graph can be considered the schematic of the overall robotic behavior. |
| Data Layer Graph | Represents the physical groupings and connections that implement the behavior modeled in the computational graph. In electrical engineering terms, if the computational graph maps to the schematic, then the data layer graph corresponds with the layout. |
| Node | A process that performs a computation. ROS is designed to be modular at a fine-grained scale; a robot control system usually comprises many nodes. Nodes execute arbitrary logic that contribute to the overall robotics behavior. |
| Topic | A continuous dataflow within a databus (the underlying communication channel) where nodes can use a publish/subscribe a communication paradigm to exchange information. Topics are generally used for continuous data streams (sensor data, robot state, etc.). They generally represent a many-to-many connection. Data might be published and subscribed at any time independent of any senders/receivers. |
| Service | On-demand blocking connection over the databus for a specific robotic task. Implemented as RPC over the computational graph. Mostly used for comparably fast tasks as requesting specific data. Semantically, used for processing requests. |
| Action | On-demand non-blocking connection over the databus generally used for any discrete behavior that moves a robot or that runs for a longer time but provides feedback during its execution. Actions can be pre-empted. Widely used to initiate lower-level control tasks. Semantically, used for real-world actions. |

***Key Message:*** *The ROS computational graph is a data structure that models the overall behavior of the robot whereas the data layer graph captures the physical grouping and connections of the robot components that implement the behavior modeled in the computational graph.*

## Past Work

Table 2 and Table 3 summarize past efforts to bring respectively ROS and ROS 2 into some of the adaptive computing platforms of Figure 6. Figure 9 depicts the most relevant contributions chronologically.

*Table 2:* **Past Research Efforts to Bring ROS into Adaptive Computing Platforms**

| Article | Year | Group |
|---|---|---|
| Proposal of ROS-compliant FPGA component for low-power robotic systems [Ref 54]:<br>Proposed for the first time, concept of ROS-compliant component aimed to simplify the integration of FPGAs into robots. Authors devise an architecture wherein hardware and software ROS components can be exchanged, however, the inner details of communicating with the programmable logic still required hardware expertise. Authors demonstrated the concept using an image processing example on a Xilinx® Zynq®-7000 SoC platform, delivering results 70% times faster with an FPGA when compared to the ordinary ROS software components running on the Arm® cores. | 2015 | Utsunomiya University |
| cReComp: Automated Design Tool for ROS-Compliant FPGA Component [Ref 55]:<br>Article proposes cReComp (creator for Reconfigurable hardware Component) as an automated design tool to improve productivity of ROS-compliant FPGA components. More specifically, cReComp is a code generator and framework for componentization in hardware using the ROS-compliant FPGA component architecture. | 2016 | Utsunomiya University |
| Architecture Exploration of Intelligent Robot System using ROS-compliant FPGA Component [Ref 56]:<br>Paper presents the exploration of a visual SLAM study case using ROS-compliant FPGA components and cReComp design tool. | 2016 | Utsunomiya University |
| Acceleration of publish/subscribe messaging in ROS-compliant FPGA component [Ref 51]:<br>Addresses the communication latency between ROS component interactions. Particularly, by using a hardware accelerated networking stack (SiTCP) and separating communications between the ROS APIs (master, slave, and parameter APIs, which use XMLRPC) and the data communications (e.g., TCPROS), the authors manage to reduce the latency of ROS nodes interacting over the network to approximately half the time (from 1ms to 0.5ms). The results are encouraging and highlight a relevant problem that must be tackled for ROS intranetwork interactions. | 2017 | Utsunomiya University |
| High Level Synthesis of ROS Protocol Interpretation and Communication Circuit for FPGA [Ref 57]:<br>Dives further into the ROS-compliant component concept and contributes to it by: a) providing a detailed description of components co-operated by the processing system (PS), b) proposed an HLS design flow for ROS-compliant component and c) studied the feasibility and evaluated the source code of an example. | 2019 | Utsunomiya University, Tokai University, and Tokyo City University |
| FPGA-ROS: Methodology to Augment the Robot Operating System with FPGA Designs [Ref 58]:<br>Work presents a proof of concept methodology to design ROS-compatible FPGA embedded systems. Authors claim support for both HDL and HLS in the methodology (though the tooling to do so is not disclosed). The main advantages of the proposed methodology include a) the automatic generation of PL-equivalents to ROS messages (standard IP to AXIS frame block), b) the chance to have a full hardware implementation of ROS entities (no need to rely on PS) and c) the future possibility of considering partial reconfiguration among the shared RTL modules. The concept is validated with an external bandwidth limited SPI-driven network module (WIZ820io). | 2019 | Technische Universität Dresden (TUD) |
| ReconfROS: Running ROS on Reconfigurable SoCs [Ref 59]: Presents an approach to integrate ROS into adaptive SoCs. Authors claim that previous implementations mostly concentrate on the development of dedicated and very specialized interface nodes with integrated preprocessing, and instead, their contribution focuses on general algorithmic acceleration. A claim to provide means for easy integration of any FPGA-based hardware accelerators into ROS nodes is made in the paper, however, the methodology to do so is not obvious to the reader. | 2021 | Osnabrück University |

*Table 3:* **Past Research Efforts to Bring ROS 2 into Adaptive Computing Platforms**

| Article | Year | Group |
| --- | --- | --- |
| ReconROS: Flexible Hardware Acceleration for ROS 2 Applications [Ref 60]: The article presents a framework that integrates ROS with ReconOS, allowing roboticists to utilize hardware acceleration for ROS applications either as hardware-accelerated ROS nodes or as ROS nodes mapped completely to hardware. | 2020 | Paderborn University, University of Klangenfurt |
| Automated Integration of High-Level Synthesis FPGA Modules with ROS2 Systems [Ref 61]: Authors present the concept of ROS 2-FPGA nodes and disclose FOrEST (FPGA-Oriented Easy Synthesizer Tool), a tool that allows an easy and seamless integration of HLS-generated FPGA logic into ROS 2 systems. The tool automatically generates ROS 2 nodes for high-level synthesis-based FPGA modules, greatly facilitating the integration of FPGAs with other robot components. FOrEST targets PYNQ 2.5, and the ROS-FPGA nodes provide a speed-up of 36.3X on simple image-related tasks. | 2020 | University of Toronto, Shibaura Institute of Technology, Keio University, and Tokai University |

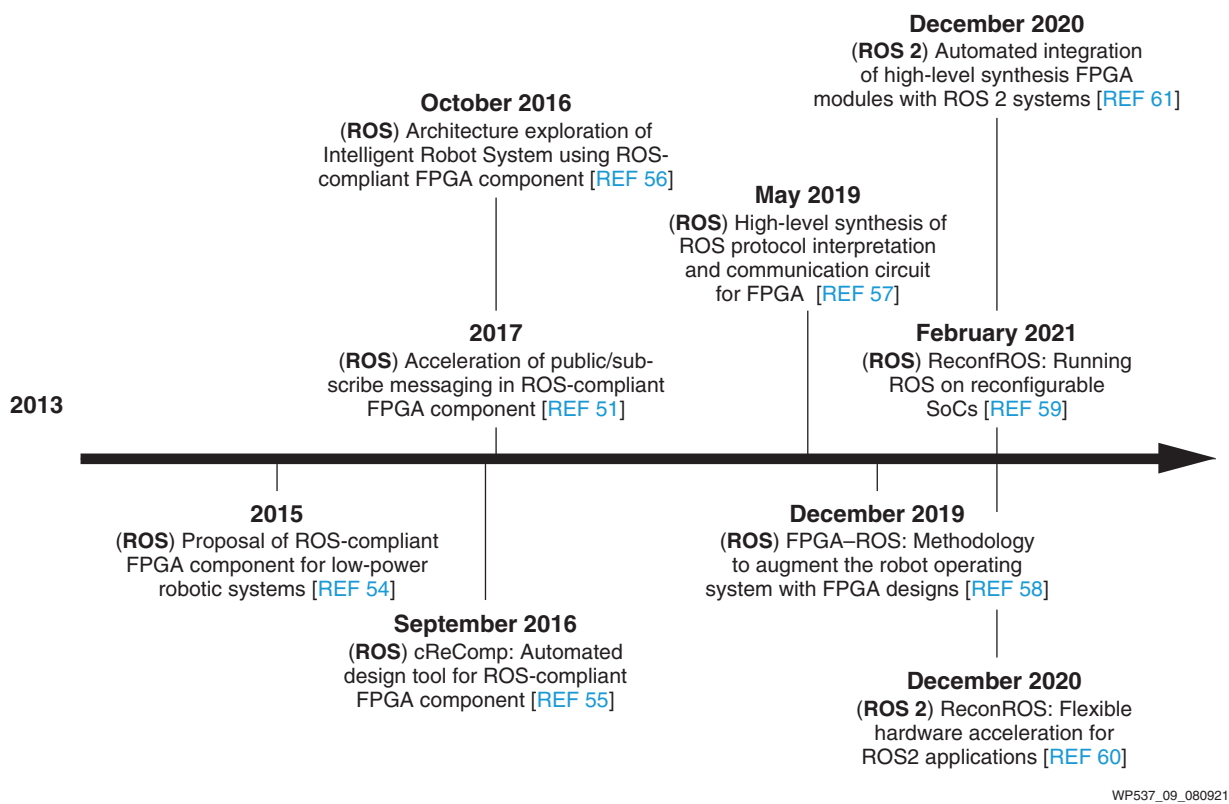

*Figure 9:* **Chronology of ROS and ROS 2 Contributions**

Figure 9 depicts the growing interest in the research community towards facilitating adaptive computing. From a ROS perspective, three groups of contributions can be identified: first, contributions like [Ref 55], [Ref 58], [Ref 59] or [Ref 61], or propose tools and methodologies that help roboticists leverage hardware acceleration capabilities and offload or accelerate parts of the ROS computational graph into the programmable logic (the FPGA). Second, contributions like [Ref 51] propose to accelerate the ROS underlayers, particularly the networking stack optimizing intra-network interactions between nodes. As described in [Ref 11], the networking stack represents a bottleneck for ROS communications, and efforts like [Ref 51] will be of relevance for real-time distributed systems. Third, examples are identified [Ref 56][Ref 62][Ref 63][Ref 64][Ref 65][Ref 66] wherein a ROS computational graph is optimized by leveraging adaptive computing.

Beyond the acceleration of particular applications and ROS libraries at the user-space level, it is worth noting the importance to accelerate interactions between ROS Nodes at the inter-process, intra-process and even at the intra-network levels. Given that robot behaviors are built as a result of ROS node interactions, accelerators on this end will impact the overall latency significantly, with relevant reductions in the overall ROS and ROS 2 computational graph dataflows. In other words, a holistic hardware acceleration view must be applied when considering ROS and ROS 2, one that accounts for a) optimizing ROS computational graph interactions inter-process, intra-process, and intra-network (including underlayers) and b) accelerating applications on top of ROS.

***Key Message:*** *A holistic hardware acceleration view must be applied when considering ROS and ROS 2, one that accounts for a) optimizing ROS computational graph interactions inter-process, intra- process and intra-network (including underlayers) and b) accelerating applications on top of ROS.*

Another observation made while reviewing the contributions in Figure 9 is that the majority of the past approaches tackle the integration of adaptive computing into ROS from a hardware engineer's perspective. The majority of the tools and methodologies proposed consider an end-user that had past experiences with embedded and hardware flows. This often implies being familiar with concepts like RTL, HDL, and HLS, or tools like Vivado® Design Suite or Vitis™ unified software platform. Similarly, deployment into embedded targets requires one to be somewhat familiar with Yocto, OpenEmbedded, and related tools. Most roboticists in the ROS world do not fit in this category. A ROS-centric approach to integrate adaptive computing is required. The hardware and embedded flows must be integrated directly into the ROS ecosystem, providing an experience similar to the one roboticists have when they build ROS workspaces in their desktop workstations. Leveraging all past work and experiences, the next section attempts this by proposing a ROS-centric architecture to integrate adaptive computing.

## ROS 2 Approach: Integrating Adaptive Computing in Robotics

Figure 10 depicts an architecture that includes hardware acceleration into the ROS 2 world while keeping a roboticist-centric view. No assumptions are made in terms of familiarity with non-ROS tools (e.g., Vivado or Vitis tools), nor with OpenEmbedded or Yocto. In addition, the architecture builds on top of open standards, focusing on C++ and OpenCL as target computing languages to generate acceleration kernels. This way, hardware acceleration capabilities become accessible to the majority in the field of robotics. The architecture builds on top of three pillars: the ROS *build system* (`ament`), the ROS *meta build tools* (`colcon`), and embedded firmware.

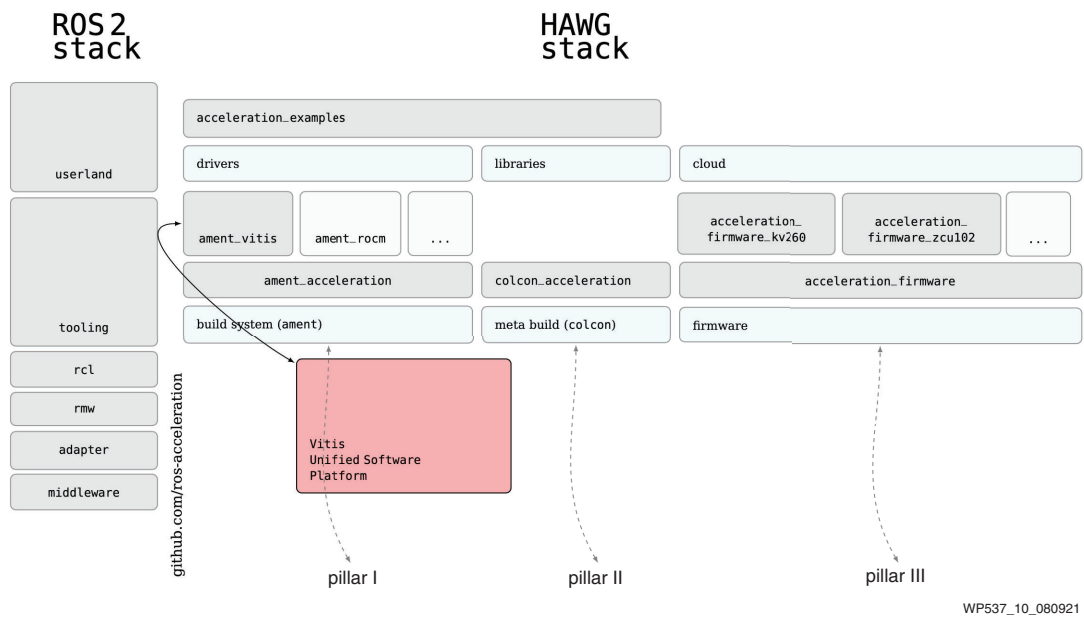

*Figure 10:* **ROS 2 Hardware Acceleration Working Group (HAWG) Initial Architecture**

The first pillar represents extensions of the ament ROS 2 build system. `ament_vitis`[1] implements such extensions through a series of CMake macros and utilities to include Vitis tools in the ROS 2 ecosystem. The proposed architecture is acceleration-technology agnostic, i.e., the `ament_acceleration` abstraction layer abstracts the build system extensions from frameworks and software platforms (e.g., Vitis tools), allowing support for FPGAs and GPUs. As an example of an alternative acceleration technology, `ament_rocm` is included in Figure 10 and illustrates the potential future integration of ROCm5[2] software development platform for HPC/Hyperscale-class GPU computing. Under the hood, each specialization of `ament_acceleration` should rely on the corresponding libraries. For example, `ament_vitis` relies on Vitis and on the Xilinx Runtime (XRT) library[3], an open-source standardized software interface that facilitates communication between the application code and the accelerated kernels. The Vitis tools and XRT are completely hidden from the robotics engineer, simplifying the creation of acceleration kernels and letting roboticists focus on improving computational graphs. This helps achieve the objective of simplifying the creation of acceleration kernels by providing an experience as if it was any other ROS package. Code listing 1 in Figure 11 shows an example using the `ament_vitis` ROS package. The macro `vitis_acceleration_kernel` gives flexibility and allows the user to seamlessly extend CMakeLists.txt files and select which parts of the ROS package to accelerate.

---

1. ament_vitis is available at https://github.com/ros-acceleration/ament_vitis

2. ROCm is available at https://github.com/RadeonOpenCompute/ROCm

3. XRT is available at https://github.com/Xilinx/XRT

**Code listing 1** ROS 2 package extension on its `CMakeLists.txt` file to generate a kernel of the type selected.

```
1  vitis_acceleration_kernel(              ← ament_vitis macro, helps define ac-
2    NAME vadd                                celerators directly from CMakeList.txt
3    FILE src/vadd.cpp                     ← kernel name
4    CONFIG src/zcu102.cfg   ← - - - - -   platform configuration
5    INCLUDE                                                          source code
6      include
7    TYPE
8      # sw_emu
9      # hw_emu                            ← build (kernel) type
10     hw
11   PACKAGE                               ← invoke Vitis compiler package flag
12 )
```

WP537_11_071521

*Figure 11:* **Code Listing 1**

The second pillar extends the `colcon` ROS meta-built tools to integrate hardware acceleration-specific flows.

The third pillar is firmware. Represented by `acceleration_firmware`[1], the third pillar is meant to provide firmware-specific artifacts for the hardware acceleration platform so that acceleration kernels can be compiled against them, simplifying the process and maintaining the ROS development flow. The proposed architecture is made purposely agnostic to the hardware acceleration platform, supporting edge (embedded) devices, PCIe® accelerators for workstations, data centers, and even cloud hardware acceleration. The selection of the platform happens by including a specific `acceleration_firmware` repository
(e.g., `acceleration_firmware_xilinx`) in the ROS workspace sources (under `src/`). The selection of the platform (in this example, it is the Xilinx Zynq UltraScale+™ MPSoC ZCU102 platform) at build time happens through `colcon mixins`. This way, `colcon build –build-base=build-zcu102 –install-base=install-zcu102 –merge-install –mixin zcu102` will build the whole ROS 2 workspace for the ZCU102 hardware platform, cross-compiling ROS packages and generating kernels as appropriate on-the-go for the ZCU102 platform. All intermediate steps are fully automated. The resulting `install-zcu102` directory can be used directly in hardware.

To allow for intermediary artifacts and provide flexibility in the embedded processes, `acceleration_firmware` introduces a new sub-folder structure into the ROS 2 workspaces, e.g., the *<ros2-workspace-path>/acceleration/firmware/<platform>* path.

Figure 12 provides a walkthough on how the architecture works when the `acceleration_examples` ROS 2 package is built. The process starts with `colcon` building the ROS 2 workspace (Figure 12, Callout 1). No special flags need to be used—only the corresponding packages and the Hardware Acceleration Working Group (HAWG) infrastructure must be in the `src` directory of the workspace. `colcon` will build each package for the development workstation architecture. This includes the `ament_vitis CMake` macros (Figure 12, Callout 2), which will deploy a series of CMake extensions into the resulting local ROS 2 overlay workspace.

---

1. `acceleration_firmware` is available at https://github.com/ros-acceleration/acceleration_firmware. See `acceleration_firmware_xilinx` for a specific example specialized for Xilinx platforms.

These extensions connect with the local Vitis installation[1] (Figure 12, Callout 3) and expose its capabilities directly to ROS 2 packages. In other words, ROS 2 packages can leverage these macros from their CMakeLists.txt files and use Xilinx's hardware acceleration tooling.

The firmware packages (e.g., `acceleration_firmware_kv260`) are a key component of the architecture. Moving from one accelerator to another should only require the replacement of this package with the one for the new target accelerator. An arbitrary valid firmware package should contain firmware artifacts and CMake logic to unpack, deploy, and configure the firmware appropriately for hardware acceleration purposes (Figure 12, Callout 4) in the ROS overlay workspace. A valid firmware package should also contain the root file system, the sysroot of the root file system (for cross-compilation) or templates to automatically generate at build-time ROS 2 `mixins`[2] to facilitate the embedded flows, among others. Refer to the complete list of artifacts in any of the official firmware packages for more details.

Altogether, a first invocation of `colcon` build (Figure 12, Callout 1) will prepare the ROS 2 workspace for hardware acceleration and deploy files in the local overlay (Figure 12, Callouts 2, 3, and 4). After that, from the overlay, a second invocation of `colcon` build with the `-mixin <target-board>` flag will cross-compile (Figure 12, Callout $5_1$)and produce the accelerators (Figure 12, Callout $5_2$) as needed for the target accelerator hardware. From this point on, the `colcon_acceleration` package helps further automate additional processes, which is beyond the scope of this white paper.

Initial support for three different boards is provided: Xilinx's Zynq UltraScale+ MPSoC ZCU102 and ZCU104 and the Kria™ KV260 Vision AI Starter Kit.

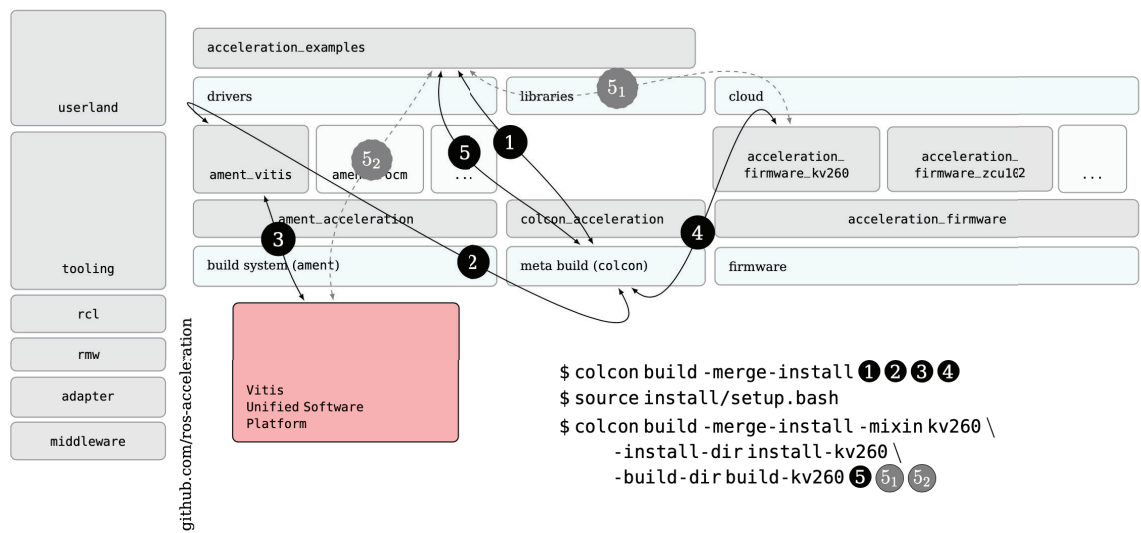

*Figure 12:* **Walkthrough on the Interaction among ROS 2 Packages in the HAWG Initial Architecture**

---

1. Vitis Unified Software Platform must be installed independently as a non-ROS dependency.

2. The mixins contain workstation-specific details (e.g. the path wherein Vitis is installed), for this reason, the mixins need to be generated dynamically and cannot be contained in an independent ROS 2 package.

**Key Message:** *Inspired by past work, the present work proposes a ROS 2-centric architecture to integrate hardware acceleration as a first class participant of the ROS ecosystem. The proposal in Figure 12 is modularized into different ROS 2 packages that can be built as any other ROS package. The architecture is platform-agnostic (i.e., considers support for edge, workstation, data center, or cloud targets), technology-agnostic (considers support for FPGAs and GPUs), and is easily portable to other boards by simply providing the corresponding* `acceleration_firmware` *specializations.*

Table 4 summarizes how the approach presented in here compares to prior efforts.

*Table 4:* **Capabilities of the Different Approaches Trying to Integrate Adaptive Computing into ROS and ROS 2, Including Xilinx's Summarized with ament_vitis**

| Capability | cReComp [Ref 51][Ref 54] [Ref 55][Ref 56] [Ref 57] | FOrEST [Ref 61] | ReconROS [Ref 60] | ReconfROS [Ref 59] | ament_vitis (this white paper) |
|---|---|---|---|---|---|
| ROS support | ✔ | | | ✔ | (possible, catkin-extensions) |
| ROS 2 support | | ✔ | ✔ | | ✔ |
| Multiple platforms/boards supported | | | | | ✔ |
| Automatic generation of PL and PS artifacts | | ✔ (partially) | | | ✔ |
| Vitis emulation targets (`sw_emu` and `hw_emu`) | | | | | ✔ |
| Declare kernels from CMakeLists.txt | | | | | ✔ |
| Integrated in ROS build system (ament) | | | | | ✔ |
| License | BSD-3-Clause | | GPL-2.0 | BSD-3-Clause | Apache 2.0 |
| Source code | kazuyamashi/cReComp | Lien182/ReconROS | ros2-forest/forest | uos/ReconfROS | ros-acceleration/ament_vitis |
| Adder example | adder.v (tutorial) | add.c (tutorial) | (other examples) | (other examples) | |

# Conclusion

Roboticists spend a significant part of their time building behaviors in the form of computational graphs that solve the robotic task at hand. They often use modern C++ to build complex real-time systems through advanced software engineering practices. However, they are not hardware engineers. Hardware and embedded expertise is scarce among roboticists, hindering the adoption of adaptive computing technologies such as FPGAs. This white paper described the foundations of adaptive computing in robotics with inspiration from past work to propose an architecture to generate ROS 2 software-defined hardware. Opposed to other past approaches, the effort presented in this white paper assumes no hardware or embedded expertise. A roboticist-centric view is used instead.

The white paper first summarized the importance of selecting the right compute platform when building a robot and the critical relationship between hardware and software in robotics. CPU, GPU, FPGA, and ASIC compute principles are briefly covered using an industrial metaphor while reviewing literature, which shows that FPGAs have clear advantages for robotics given their low power, high performance, deterministic, reconfigurable, secure, and adaptive characteristics. Past work integrating ROS into FPGA platforms is also considered, leading to the observation that these approaches use a hardware engineer's perspective and that the resulting design decisions often significantly limit the outcome possibilities. Against this, a ROS 2 roboticist's perspective is proposed, one that holistically considers a hardware acceleration view for ROS 2 that accounts for a) optimizing ROS 2 computational graph interactions inter-process, intra-process, and intra-network (including underlayers) and b) seamlessly accelerates applications on top of ROS 2.

An architecture to realize this objective is proposed in the form of a set of ROS 2 packages, which deliver hardware acceleration into the ROS workspace. The architecture is platform-agnostic (i.e., considers support for edge, workstation, data center, or cloud targets), technology-agnostic (considers support for FPGAs and GPUs), and easily portable to other boards. The core components of the architecture are disclosed under an Apache 2.0 license, and initial support for three boards is demonstrated. The architecture provided is also application-agnostic and can easily be ported to ROS in the future by including `catkin` extensions (as opposed to `ament`).

For additional information on Xilinx Robotics, go to
https://www.xilinx.com/applications/industrial/robotics.html and
https://github.com/ros-acceleration

# References

1. V. Mayoral-Vilches, A. Hernández, R. Kojcev, I. Muguruza, I. Zamalloa, A. Bilbao, and L. Usategi, "The shift in the robotics paradigm-the hardware robot operating system (h-ros); an infrastructure to create interoperable robot components," in 2017 NASA/ESA Conference on Adaptive Hard- ware and Systems (AHS). IEEE, 2017, pp. 229-236.

2. L. A. Kirschgens, I. Z. Ugarte, E. G. Uriarte, A. M. Rosas, and V. Mayoral-Vilches, "Robot hazards: from safety to security," arXiv preprint arXiv:1806.06681, 2018.

3. Z. Wan, B. Yu, T. Y. Li, J. Tang, Y. Zhu, Y. Wang, A. Raychowdhury, and S. Liu, "A survey of FPGA-based robotic computing," IEEE Circuits and Systems Magazine, vol. 21, no. 2, pp. 48-74, 2021.

4. V. Mayoral-Vilches, L. U. S. Juan, B. Dieber, U. A. Carbajo, and E. Gil-Uriarte, "Introducing the robot vulnerability database (rvd)," arXiv preprint arXiv:1912.11299, 2019.

5. G. Lacava, A. Marotta, F. Martinelli, A. Saracino, A. La Marra, E. Gil-Uriarte, and V. Mayoral-Vilches, "Current research issues on cyber security in robotics," 2020.

6. Q. Zhu, S. Rass, B. Dieber, and V. Mayoral-Vilches, "Cybersecurity in robotics: Challenges, quantitative modeling, and practice," arXiv preprint arXiv:2103.05789, 2021.

7. V. Mayoral-Vilches, N. García-Maestro, M. Towers, and E. Gil-Uriarte, "Devsecops in robotics," arXiv preprint arXiv:2003.10402, 2020.

8. V. Mayoral-Vilches, L. A. Kirschgens, A. B. Calvo, A. H. Cordero, R. I. Pisón, D. M. Vilches, A. M. Rosas, G. O. Mendia, L. U. S. Juan, I. Z. Ugarte et al., "Introducing the robot security framework (rsf), a standardized methodology to perform security assessments in robotics," arXiv preprint arXiv:1806.04042, 2018.

9. V. Mayoral-Vilches, E. Gil-Uriarte, I. Zamalloa Ugarte, G. Olalde Mendia, R. Izquierdo Pisón, L. Alzola Kirschgens, A. Bilbao Calvo, A. Hernández Cordero, L. Apa, and C. Cerrudo, "Towards an open standard for assessing the severity of robot security vulnerabilities, the robot vulnerability scoring system (rvss)," arXiv preprint arXiv:1807.10357, 2018.

10. C. S. V. Gutiérrez, L. U. S. Juan, I. Z. Ugarte, and V. Mayoral-Vilches, "Time-sensitive networking for robotics," arXiv preprint arXiv:1804.07643, 2018.

11. C. S. V. Gutiérrez, L. U. S. Juan, I. Z. Ugarte, and V. Mayoral-Vilches, "Real-time Linux communications: an evaluation of the linux communication stack for real- time robotic applications," arXiv preprint arXiv:1808.10821, 2018.

12. C. S. V. Gutiérrez, L. U. S. Juan, I. Z. Ugarte, and V. Mayoral-Vilches, "Towards a distributed and real-time framework for robots: Evaluation of ROS 2.0 communi- cations for real-time robotic applications," arXiv preprint arXiv:1809.02595, 2018.

13. C. S. V. Gutiérrez, L. U. S. Juan, I. Z. Ugarte, I. M. Goenaga, L. A. Kirschgens, and V. Mayoral-Vilches, "Time synchronization in modular collaborative robots," arXiv preprint arXiv:1809.07295, 2018.

14. V. Mayoral-Vilches, R. Kojcev, N. Etxezarreta, A. Hernández, and I. Zamalloa, "Towards self-adaptable robots: from programming to training machines," arXiv preprint arXiv:1802.04082, 2018.

15. W. Lie and W. Feng-Yan, "Dynamic partial reconfiguration in FPGAS," in 2009 Third International Symposium on Intelligent Information Technology Application, vol. 2. IEEE, 2009, pp. 445-448.

16. R. Brooks, "A robust layered control system for a mobile robot," IEEE journal on robotics and automation, vol. 2, no. 1, pp. 14-23, 1986.

17. R. A. Brooks and J. H. Connell, "Asynchronous distributed control system for a mobile robot," in Mobile Robots I, vol. 727. International Society for Optics and Photonics, 1987, pp. 77-84.

18. T. Ziemke, "The construction of 'reality' in the robot: Constructivist perspectives on situated artificial intelligence and adaptive robotics," Foundations of science, vol. 6, no. 1, pp. 163-233, 2001.

19. J. Willmann, P. Block, M. Hutter, K. Byrne, and T. Schork, Robotic Fabrication in Architecture, Art and Design 2018: Foreword by Sigrid Brell-Çokcan and Johannes Braumann, Association for Robots in Architecture. Springer, 2018.

20. V. Mayoral-Vilches, R. Kojcev, N. Etxezarreta, A. Hernández, and I. Zamalloa, "Towards self-adaptable robots: from programming to training machines," arXiv preprint arXiv:1802.04082, 2018.

21. V. Mayoral-Vilches, R. Kojcev, A. Hernández, I. Zamalloa, and A. Bilbao, "Modular and self-adaptable (masa) strategy for building robots," in 2018 NASA/ESA Conference on Adaptive Hard- ware and Systems (AHS). IEEE, 2018, pp. 90-95.

22. R. Kojcev, N. Etxezarreta, A. Hernández, and V. Mayoral-Vilches, "Hierarchical learning for modu- lar robots," arXiv preprint arXiv:1802.04132, 2018.

23. C. M. Gosselin, "Adaptive robotic mechanical systems: A design paradigm," 2006.

24. M. T. Mason, J. K. Salisbury, and J. K. Parker, "Robot hands and the mechanics of manipulation," 1989.

25. Butterfaß, M. Grebenstein, H. Liu, and G. Hirzinger, "Dlr-hand ii: Next generation of a dex- trous robot hand," in Proceedings 2001 ICRA. IEEE International Conference on Robotics and Automation (Cat. No. 01CH37164), vol. 1. IEEE, 2001, pp. 109-114.

26. F. Y. Chen, "Gripping mechanisms for industrial robots: an overview," Mechanism and Machine Theory, vol. 17, no. 5, pp. 299-311, 1982.

27. K. S. Ivanov, "Adaptive robotics," in Applied Mechanics and Materials, vol. 656. Trans Tech Publ, 2014, pp. 154-163.

28. A. C. Sanderson and L. E. Weiss, "Adaptive visual servo control of robots," in Robot vision. Springer, 1983, pp. 107-116.

29. R. Carelli, O. Nasisi, and B. Kuchen, "Adaptive robot control with visual feedback," in Proceedings of 1994 American Control Conference-ACC'94, vol. 2. IEEE, 1994, pp. 1757-1760.

30. J. Lee, "Apply force/torque sensors to robotic applications," Robotics, vol. 3, no. 2, pp. 189-194, 1987.

31. S. Lu, J. H. Chung, and S. A. Velinsky, "Human-robot collision detection and identification based on wrist and base force/torque sensors," in Proceedings of the 2005 IEEE International Conference on Robotics and Automation. IEEE, 2005, pp. 3796-3801.

32. J. Henkel and L. Bauer, "What is adaptive computing?" SIGDA Newsl., vol. 40, no. 5, p. 1, May 2010. [Online]. Available: https://doi.org/10.1145/1866966.1866967.

33. S. Murray, W. Floyd-Jones, Y. Qi, D. J. Sorin, and G. D. Konidaris, "Robot motion planning on a chip," in Robotics: Science and Systems, 2016.

34. S. Murray, W. Floyd-Jones, G. Konidaris, and D. J. Sorin, "A programmable architecture for robot motion planning acceleration," in 2019 IEEE 30th International Conference on Application-specific Systems, Architectures and Processors (ASAP), vol. 2160. IEEE, 2019, pp. 185-188.

35. B. Plancher, S. M. Neuman, T. Bourgeat, S. Kuindersma, S. Devadas, and V. J. Reddi, "Accelerating robot dynamics gradients on a CPU, GPU, and FPGA," IEEE Robotics and Automation Letters, vol. 6, no. 2, pp. 2335-2342, 2021.

36. S. M. Neuman, B. Plancher, T. Bourgeat, T. Tambe, S. Devadas, and V. J. Reddi, "Robomorphic computing: a design methodology for domain-specific accelerators parameterized by robot morphology," in Proceedings of the 26th ACM International Conference on Architectural Support for Programming Languages and Operating Systems, 2021, pp. 674-686.

37. S. M. Neuman, "A design methodology for computer architecture parameterized by robot morphology," Ph.D. dissertation, Massachusetts Institute of Technology, 2020.

38. K. Sugiura and H. Matsutani, "An FPGA acceleration and optimization techniques for 2d LiDAR slam algorithm," IEICE TRANSACTIONS on Information and Systems, vol. 104, no. 6, pp. 789-800, 2021.

39. B. Williams, "Evaluation of a SoC for real-time 3d slam," 2017.
40. R. Freeman,"Configurable electrical circuit having configurable logic elements and configurable interconnects," Sep 1989, US Patent 4,870,302. [Online]. Available: https://patents.google.com/patent/US4870302A
41. S. Murray, W. Floyd-Jones, Y. Qi, G. Konidaris, and D. J. Sorin, "The microarchitecture of a real-time robot motion planning accelerator," in 2016 49th Annual IEEE/ACM International Symposium on Microarchitecture (MICRO). IEEE, 2016, pp. 1-12.
42. G. J. García, C. A. Jara, J. Pomares, A. Alabdo, L. M. Poggi, and F. Torres, "A survey on FPGA-based sensor systems: towards intelligent and reconfigurable low-power sensors for computer vision, control and signal processing," Sensors, vol. 14, no. 4, pp. 6247-6278, 2014.
43. J. J. Rodríguez-Andina, M. D. Valdes-Pena, and M. J. Moure, "Advanced features and industrial applications of FPGAs—a review," IEEE Transactions on Industrial Informatics, vol. 11, no. 4, pp. 853-864, 2015.
44. M. Lipp, M. Schwarz, D. Gruss, T. Prescher, W. Haas, S. Mangard, P. Kocher, D. Genkin, Y. Yarom, and M. Hamburg, "Meltdown," arXiv preprint arXiv:1801.01207, 2018.
45. P. Kocher, J. Horn, A. Fogh, D. Genkin, D. Gruss, W. Haas, M. Hamburg, M. Lipp, S. Mangard, T. Prescher et al., "Spectre attacks: Exploiting speculative execution," in 2019 IEEE Symposium on Security and Privacy (SP). IEEE, 2019, pp. 1-19.
46. Xilinx, "Vitis™ unified software platform documentation - application acceleration development," UG1393 (v2020.2) March 22, 2021, 2021. [Online]. Available: https://www.xilinx.com/support/ documentation/sw_manuals/xilinx2020_2/ug1393-vitis-application-acceleration.pdf
47. J. Von Neumann, "First draft of a report on the edvac," IEEE Annals of the History of Computing, vol. 15, no. 4, pp. 27-75, 1993.
48. G. Martin, "Adaptive computing, technology overview," Xilinx, 2021.
49. A. Kosuge, K. Yamamoto, Y. Akamine, and T. Oshima, "An SoC-FPGA-based iterative-closest-point accelerator enabling faster picking robots," IEEE Transactions on Industrial Electronics, vol. 68, no. 4, pp. 3567-3576, 2020.
50. N. Frick, "Advantages of FPGA based robot control compared to CPU and MCU based control methods," 2020.
51. Y. Sugata, T. Ohkawa, K. Ootsu, and T. Yokota, "Acceleration of publish/subscribe messaging in ROS-compliant FPGA component," in Proceedings of the 8th International Symposium on Highly Efficient Accelerators and Reconfigurable Technologies, 2017, pp. 1-6.
52. H. Cheng, S. Sato, and H. Nakahara, "A performance per power efficient object detector on an FPGA for robot operating system (ROS)," in Proceedings of the 9th International Symposium on Highly-Efficient Accelerators and Reconfigurable Technologies, 2018, pp. 1-4.
53. M. Quigley, K. Conley, B. Gerkey, J. Faust, T. Foote, J. Leibs, R. Wheeler, and A. Y. Ng, "ROS: an open-source robot operating system," in ICRA workshop on open source software, vol. 3, no. 3.2. Kobe, Japan, 2009, p. 5.
54. K. Yamashina, T. Ohkawa, K. Ootsu, and T. Yokota, "Proposal of ROS-compliant FPGA component for low-power robotic systems," arXiv preprint arXiv:1508.07123, 2015.
55. K. Yamashina, H. Kimura, T. Ohkawa, K. Ootsu, and T. Yokota, "cReComp: Automated design tool for ROS-compliant FPGA component," in 2016 IEEE 10th International Symposium on Embedded Multicore/Many-core Systems-on-Chip (MCSOC). IEEE, 2016, pp. 138-145.
56. T. Ohkawa, K. Yamashina, T. Matsumoto, K. Ootsu, and T. Yokota, "Architecture exploration of intelligent robot system using ROS-compliant FPGA component," in 2016 International Symposium on Rapid System Prototyping (RSP). IEEE, 2016, pp. 1-7.

**XILINX** Adaptive Computing in Robotics Leveraging ROS 2 to Enable Software-Defined Hardware for FPGAs57. T. Ohkawa, Y. Sugata, H. Watanabe, N. Ogura, K. Ootsu, and T. Yokota, "High level synthesis of ROS protocol interpretation and communication circuit for FPGA," in 2019 IEEE/ACM 2nd International Workshop on Robotics Software Engineering (RoSE). IEEE, 2019, pp. 33-36.

58. A. Podlubne and D. Göhringer, "Fpga-ros: Methodology to augment the robot operating system with fpga designs," in 2019 International Conference on ReConFigurable Computing and FPGAs (ReConFig). IEEE, 2019, pp. 1-5.

59. M. Eisoldt, S. Hinderink, M. Tassemeier, M. Flottmann, J. Vana, T. Wiemann, J. Gaal, M. Rothmann, and M. Porrmann, "Reconfros: Running ROS on reconfigurable SoCs," in Proceedings of the 2021 Drone Systems Engineering and Rapid Simulation and Performance Evaluation: Methods and Tools Proceedings, 2021, pp. 16-21.

60. C. Lienen, M. Platzner, and B. Rinner, "Reconros: Flexible hardware acceleration for ROS2 applications," in 2020 International Conference on Field-Programmable Technology (ICFPT). IEEE, 2020, pp. 268-276.

61. D. P. Leal, M. Sugaya, H. Amano, and T. Ohkawa, "Automated integration of high-level synthesis FPGA modules with ROS 2 systems," in 2020 International Conference on Field-Programmable Technology (ICFPT), 2020, pp. 292-293.

62. S. Panadda, J. Nattha, P. L. Daniel, and O. Takeshi, "Low-power high-performance intelligent camera framework ROS-FPGA node," in Proceedings of Asia Pacific Conference on Robot IoT System Development and Platform, no. 2020, 2021, pp. 73-74.

63. J. P. Queralta, F. Yuhong, L. Salomaa, L. Qingqing, T. N. Gia, Z. Zou, H. Tenhunen, and T. Westerlund, "FPGA-based architecture for a low-cost 3d lidar design and implementation from multiple rotating 2d lidars with ros," in 2019 IEEE SENSORS, 2019, pp. 1-4.

64. T. K. Maiti, "Ros on arm processor embedded with fpga for improvement of robotic computing," in 2021 International Symposium on Devices, Circuits and Systems (ISDCS), 2021, pp. 1-4.

65. T. Ohkawa, K. Yamashina, H. Kimura, K. Ootsu, and T. Yokota, "FPGA components for integrating fpgas into robot systems," IEICE TRANSACTIONS on Information and Systems, vol. 101, no. 2, pp. 363-375, 2018.

66. D. P. Leal, M. Sugaya, H. Amano, and T. Ohkawa, "FPGA acceleration of ROS2-based reinforcement learning agents," in 2020 Eighth International Symposium on Computing and Networking Workshops (CANDARW), 2020, pp. 106-112.
WP537 (v1.0) www.xilinx.com 26

# Revision History

The following table shows the revision history for this document:

| Date | Version | Description of Revisions |
|---|---|---|
| 08/25/2021 | 1.0 | Initial Xilinx release. |

# Disclaimer

# Automotive Applications Disclaimer